\documentclass{amia}
\usepackage{graphicx}
\usepackage[labelfont=bf]{caption}
\usepackage[superscript,nomove]{cite}
\usepackage{color}
\usepackage{amssymb}
\usepackage{bibspacing}
\usepackage{multicol}

\begin{document}

\title{NRC-Canada at SMM4H Shared Task: Classifying Tweets Mentioning Adverse Drug Reactions and Medication Intake}

\author{Svetlana Kiritchenko, Ph.D., Saif M. Mohammad, Ph.D., Jason Morin, JCD, Ph.D.,\\ \vspace{1mm} Berry de Bruijn, Ph.D.}

\institutes{
    \vspace{1mm} National Research Council Canada, Ottawa, ON, Canada\\
	\vspace{2mm}{\tt \small \{svetlana.kiritchenko,saif.mohammad,jason.morin,berry.debruijn\}@nrc-cnrc.gc.ca}\\
}
\maketitle

\noindent{\bf Abstract}

\textit{Our team, NRC-Canada, participated in two shared tasks at the AMIA-2017 Workshop on Social Media Mining for Health Applications (SMM4H): Task 1 - classification of tweets mentioning adverse drug reactions, and Task 2 - classification of tweets describing personal medication intake. For both tasks, we trained Support Vector Machine classifiers using a variety of surface-form, sentiment, and domain-specific features. With nine teams participating in each task, our submissions ranked first on Task 1 and third on Task 2. Handling considerable class imbalance proved crucial for Task 1. We applied an under-sampling technique to reduce class imbalance (from about 1:10 to 1:2). Standard $n$-gram features, $n$-grams generalized over domain terms, as well as general-domain and domain-specific word embeddings had a substantial impact on the overall performance in both tasks. On the other hand, including sentiment lexicon features did not result in any improvement.}

\section{Introduction}
Adverse drug reactions (ADR)---unwanted or harmful reactions resulting from correct medical drug use---present a significant and costly public health problem.\cite{lazarou1998incidence} 
Detecting, assessing, and preventing these events are the tasks of {\it pharmacovigilance}. 
In the pre-trial and trial stages of drug development, the number of people taking a drug is carefully controlled, and the collection of ADR data is centralized. However, after the drug is available widely, post-marketing surveillance often requires the collection and merging of data from disparate sources,\cite{Waller2017} including patient-initiated spontaneous reporting.
Unfortunately, adverse reactions to drugs are grossly underreported to health professionals.\cite{pmid15141997, pmid28104709}  
Considerable issues with patient-initiated reporting have been identified, including various types of reporting biases and causal attributions of adverse events.\cite{pmid28694146, pmid24312892, pmid26163365}  
Nevertheless, a large number of people, freely and spontaneously, report ADRs on social media. The potential availability of inexpensive, large-scale, and real-time data on ADRs makes social media a valuable resource for pharmacovigilance. 

Information required for pharmacovigilance includes a reported adverse drug reaction, a linked drug referred to by its full, abbreviated, or generic name, and an indication whether it was the social media post author that experienced the adverse event. 
However, there are considerable challenges in automatically extracting this information from free-text social media data. 
Social media texts are often short and informal, and include non-standard abbreviations and creative language. 
Drug names or their effects may be mis-spelled; they may be used metaphorically
(e.g., \textit{Physics is like higher level maths on steroids}). Drug names might have other non-drug related meanings (e.g., {\it ecstasy}). An adverse event may be negated or only expected (e.g., \textit{I bet I'll be running to the bathroom all night}), or it may not apply to the author of the post at all (e.g., a re-tweet of a press release).

The shared task challenge organized as part of the AMIA-2017 Workshop on Social Media Mining for Health Applications (SMM4H) focused on Twitter data and had three tasks: 
Task 1 - recognizing whether a tweet is reporting an adverse drug reaction, 
Task 2 - inferring whether a tweet is reporting the intake of a medication by the tweeter, 
and Task 3 - mapping a free-text ADR to a standardized MEDDRA term.  
Our team made submissions for Task 1 and Task 2. For both tasks, we trained Support Vector Machine classifiers using a variety of surface-form, sentiment, and domain-specific features.  Handling class imbalance with under-sampling was particularly helpful. 
Our submissions obtained F-scores of 0.435 on Task 1 and 0.673 on Task 2, resulting in a rank of first and third, respectively. (Nine teams participated in each task.)
We make the resources created as part of this project freely available at the project webpage: http://saifmohammad.com/WebPages/tweets4health.htm. 

\newpage
\section{Task and Data Description}
\label{data}

Below we describe in detail the two tasks we participated in, Task 1 and Task 2.

\vspace{2mm}
{\bf Task 1: Classification of Tweets for Adverse Drug Reaction}

\setlength{\tabcolsep}{12pt}

\begin{table*}[t!]
\begin{center}
\begin{tabular}{llrrr}
\hline {\bf Dataset} & & {\bf Class 1 (ADR)} & {\bf Class 0 (non-ADR)} & {\bf All}\\ 
\hline
Training set	& \ \ \ \ \ & 732 (12\%) & 5,519 (88\%) & 6,251\\
Development	set & & 241 (7\%) & 3,302 (93\%) & 3,543\\
Test set	& & 771 (8\%) &  9,190 (92\%) & 9,961\\
\hline
\end{tabular}
\caption{The number of available instances in the training, development, and test sets for Task 1.}
\label{tab:task1-data-stat}
\end{center}
\end{table*}

Task 1 was formulated as follows: given a tweet, determine whether it mentions an adverse drug reaction. 
This was a binary classification task:
\begin{itemize}
\item class 1 (ADR) - tweets that mention adverse drug reactions\\
\textit{Example: Nicotine lozenges are giving me stomach cramps.}
\item class 0 (non-ADR) - tweets that do not mention adverse drug reactions\\
\textit{Example: I need a injection of Prozac ! ...now!!!!}
\end{itemize}
The official evaluation metric was the F-score for class 1 (ADR): 
\[P_{class\:1} = \frac{TP_{class\:1}}{TP_{class\:1} + FP_{class\:1}},\ \ \ \ \ R_{class\:1} = \frac{TP_{class\:1}}{TP_{class\:1} + FN_{class\:1}},\ \ \ \ \ F_{class\:1} = \frac{2 \times P_{class\:1} \times R_{class\:1}}{P_{class\:1} + R_{class\:1}}\]

The data for this task was created as part of a large project on ADR detection from social media by the DIEGO lab at Arizona State University. 
The tweets were collected using the generic and brand names of the drugs as well as their phonetic misspellings. 
Two domain experts under the guidance of a pharmacology expert annotated the tweets for the presence or absence of an ADR mention. 
The inter-annotator agreement for the two annotators was Cohens Kappa $\kappa = 0.69$.\cite{SMM4H2016} 

Two labeled datasets were provided to the participants: a training set containing 10,822 tweets and a development set containing 4,845 tweets. 
These datasets were distributed as lists of tweet IDs, and the participants needed to download the tweets using the provided Python script. 
However, only about 60--70\% of the tweets were accessible at the time of download (May 2017). 
The training set contained several hundreds of duplicate or near-duplicate messages, which we decided to remove. 
Near-duplicates were defined as tweets containing mostly the same text but differing in user mentions, punctuation, or other non-essential context. 
A separate test set of 9,961 tweets was provided without labels at the evaluation period. 
This set was distributed to the participants, in full, by email. 
Table~\ref{tab:task1-data-stat} shows the number of instances we used for training and testing our model.  

Task 1 was a rerun of the shared task organized in 2016.\cite{SMM4H2016} 
The best result obtained in 2016 was $F_{class\:1} = 0.42$.\cite{MayoNLP2016}  
The participants in the 2016 challenge employed various statistical machine learning techniques, such as Support Vector Machines, Maximum Entropy classifiers, Random Forests, and other ensembles.\cite{MayoNLP2016,SwissChocolate2016}  
A variety of features (e.g., word $n$-grams, word embeddings, sentiment, and topic models) as well as extensive medical resources (e.g., UMLS, lexicons of ADRs, drug lists, and lists of known drug-side effect pairs) were explored.

\vspace{2mm}
{\bf Task 2: Classification of Tweets for Medication Intake}

Task 2 was formulated as follows: given a tweet, determine if it mentions personal medication intake, possible medication intake, or no intake is mentioned. 
This was a multi-class classification problem with three classes:
\begin{itemize}
\vspace*{-2mm}
\item class 1 (personal medication intake) - tweets in which the user clearly expresses a personal medication intake/consumption\\
\textit{Example: Advil just saved my life :))}
\vspace*{-2mm}
\item class 2 (possible medication intake) - tweets that are ambiguous but suggest that the user may have taken the medication\\
\textit{Example: Having pains and all my Tylenol gone}
\vspace*{-2mm}
\item class 3 (non-intake) - tweets that mention medication names but do not indicate personal intake\\
\textit{Example: Going thru this pain without Tylenol..} 
\end{itemize}

The official evaluation metric for this task was micro-averaged F-score of the class 1 (intake) and class 2 (possible intake):
\[P_{class\:1\:+\:class\:2} = \frac{TP_{class\:1} + TP_{class\:2}}{TP_{class\:1} + FP_{class\:1} + TP_{class\:2} + FP_{class\:2}}\]

\[R_{class\:1 + class\:2} = \frac{TP_{class\:1} + TP_{class\:2}}{TP_{class\:1} + FN_{class\:1} + TP_{class\:2} + FN_{class\:2}}\]

\[F_{class\:1 + class\:2} = \frac{2 \times P_{class\:1 + class\:2} \times R_{class\:1 + class\:2}}{P_{class\:1 + class\:2} + R_{class\:1 + class\:2}}\]

\begin{table*}[t!]
\begin{center}
\begin{tabular}{llrrrr}
\hline {\bf Dataset} &	& {\bf Class 1} & {\bf Class 2} & {\bf Class 3} & {\bf All}\\
& & {\bf (intake)} & {\bf (possible intake)} & {\bf (non-intake)}\\
\hline
Training set	&  \ \ \ \ \ & 1,475 (20\%) & 2,374 (31\%) & 3,679 (49\%) & 7,528\\
Development	set & & 398 (19\%) & 664 (32\%) & 1,006 (49\%) & 2,068\\
Test set	& & 1,731 (23\%) & 2,697 (36\%) & 3,085 (41\%) & 7,513\\
\hline
\end{tabular}
\caption{The number of available instances in the training, development, and test sets for Task 2.}
\label{tab:task2-data-stat}
\end{center}
\end{table*}

\setlength{\tabcolsep}{6pt}

Information on how the data was collected and annotated was not available until after the evaluation.

Two labeled datasets were provided to the participants: a training set containing 8,000 tweets and a development set containing 2,260 tweets. 
As for Task 1, the training and development sets were distributed through tweet IDs and a download script. 
Around 95\% of the tweets were accessible through download. 
Again, we removed duplicate and near-duplicate messages. 
A separate test set of 7,513 tweets was provided without labels at the evaluation period. 
This set was distributed to the participants, in full, by email. 
Table~\ref{tab:task2-data-stat} shows the number of instances we used for training and testing our model.

For each task, three submissions were allowed from each participating team.

\section{System Description}
\label{system}

Both our systems, for Task 1 and Task 2, share the same classification framework and feature pool. 
The specific configurations of features and parameters were chosen for each task separately through cross-validation experiments (see Section~\ref{submissions}). 

\subsection{Machine Learning Framework}

For both tasks, we trained linear-kernel Support Vector Machine (SVM) classifiers. 
Past work has shown that SVMs are effective on text categorization tasks and robust when working with large feature spaces. 
In our cross-validation experiments on the training data, a linear-kernel SVM trained with the features described below was able to obtain better performance than a number of other statistical machine-learning algorithms, such as Stochastic Gradient Descent, AdaBoost, Random Forests, as well SVMs with other kernels (e.g., RBF, polynomic). 
We used an in-house implementation of SVM. 

{\bf Handling Class Imbalance: } For Task 1 (Classification of tweets for ADR), the provided datasets were highly imbalanced: the ADR class occurred in less than 12\% of instances in the training set and less than 8\% in the development and test sets. 
Most conventional machine-learning algorithms experience difficulty with such data, classifying  most of the instances into the majority class. 
Several techniques have been proposed to address the issue of class imbalance, including over-sampling, under-sampling, cost-sensitive learning, and ensembles.\cite{haixiang2017learning} 
We experimented with several such techniques. 
The best performance in our cross-validation experiments 
was obtained using under-sampling with the class proportion 1:2. 
To train the model, we provided the classifier with all available data for the minority class (ADR) and a randomly sampled subset of the majority class (non-ADR) data in such a way that the number of instances in the majority class was twice the number of instances in the minority class. 
We found that this strategy significantly outperformed the more traditional balanced under-sampling where the majority class is sub-sampled to create a balanced class distribution. 
In one of our submissions for Task 1 (submission 3), we created an ensemble of three classifiers trained on the full set of instances in the minority class (ADR) and different subsets of the majority class (non-ADR) data. 
We varied the proportion of the majority class instances to the minority class instances: 1:2, 1:3, and 1:4. 
The final predictions were obtained by majority voting on the predictions of the three individual classifiers. 

For Task 2 (Classification of tweets for medication intake), the provided datasets were also imbalanced but not as much as for Task 1: the class proportion in all subsets was close to 1:2:3. 
However, even for this task, we found some of the techniques for reducing class imbalance helpful. 
In particular, training an SVM classifier with different class weights improved the performance in the cross-validation experiments. 
These class weights are used to increase the cost of misclassification errors for the corresponding classes. 
The cost for a class is calculated as the generic cost parameter (parameter C in SVM) multiplied by the class weight.  
The best performance on the training data was achieved with class weights set to 4 for class 1 (intake), 2 for class 2 (possible intake), and 1 for class 3 (non-intake).

{\bf Preprocessing:} The following pre-processing steps were performed.
URLs and user mentions were normalized to http://someurl and @username, respectively. 
Tweets were tokenized with the CMU Twitter NLP tool.\cite{Gimpel11}

\subsection{Features}

The classification model leverages a variety of general textual features as well as sentiment and domain-specific features described below.
Many features were inspired by previous work on ADR \cite{sarker2015portable,MayoNLP2016,SwissChocolate2016} and our work on sentiment analysis (such as the winning system in the SemEval-2013 task on sentiment analysis in Twitter \cite{MohammadSemEval2013} and best performing stance detection system \cite{MohammadSK17}).

\vspace*{2mm}
\textit{General Textual Features}

The following surface-form features were used:

\vspace*{-3mm}
\begin{itemize}
\item $N$-grams: word $n$-grams (contiguous sequences of $n$ tokens), non-contiguous word $n$-grams ($n$-grams with one token replaced by *),
character $n$-grams (contiguous sequences of $n$ characters), unigram stems obtained with the Porter stemming algorithm;
\vspace*{-2mm}
\item General-domain word embeddings: 
\begin{itemize}
\vspace*{-2mm}
\item dense word representations generated with word2vec on ten million English-language tweets, summed over all tokens in the tweet,
\vspace*{-1mm}
\item word embeddings distributed as part of ConceptNet 5.5\cite{speer2017conceptnet}, summed over all tokens in the tweet;
\end{itemize} 
\vspace*{-3mm}
\item General-domain word clusters: presence of tokens from the word clusters generated with the Brown clustering algorithm on 56 million English-language tweets;\cite{Gimpel11} 
\vspace*{-2mm}
\item Negation: presence of simple negators (e.g., \textit{not}, \textit{never}); negation also affects the $n$-gram features---a term $t$ becomes $t\_NEG$ if it occurs after a negator and before a punctuation mark;
\vspace*{-2mm}
\item Twitter-specific features: the number of tokens with all characters in upper case, the number of hashtags, presence of positive and negative emoticons, whether the last token is a positive or negative emoticon, the number of elongated words (e.g.,  \textit{soooo});
\vspace*{-2mm}
\item Punctuation: presence of exclamation and question marks, whether the last token contains an exclamation or question mark.
\end{itemize}

\textit{Domain-Specific Features}

To generate domain-specific features, we used the following domain resources:
\begin{itemize}
\vspace*{-2mm}
\item Medication list: we compiled a medication list by selecting all one-word medication names from RxNorm (e.g, \textit{acetaminophen}, \textit{nicorette}, \textit{zoloft}) since most of the medications mentioned in the training datasets were one-word strings.  
\vspace*{-2mm}
\item Pronoun Lexicon: we compiled a lexicon of first-person pronouns (e.g., \textit{I}, \textit{ours}, \textit{we'll}), second-person pronouns (e.g., \textit{you}, \textit{yourself}), and third-person pronouns (e.g., \textit{them}, \textit{mom's}, \textit{parents'}).  
\vspace*{-2mm}
\item ADR Lexicon: a list of 13,699 ADR concepts compiled from COSTART, SIDER, CHV, and drug-related tweets by the DIEGO lab;\cite{nikfarjam2015pharmacovigilance}
\vspace*{-2mm}
\item domain word embeddings: dense word representations generated by the DIEGO lab by applying word2vec on one million tweets mentioning  medications;\cite{nikfarjam2015pharmacovigilance}
\vspace*{-2mm}
\item domain word clusters: word clusters generated by the DIEGO lab using the word2vec tool to perform
K-means clustering on the above mentioned domain word embeddings.\cite{nikfarjam2015pharmacovigilance} 
\end{itemize}

From these resources, the following domain-specific features were generated:

\vspace*{-3mm}
\begin{itemize}
\item $N$-grams generalized over domain terms (or domain generalized $n$-grams, for short): $n$-grams where words or phrases representing a medication (from our medication list) or an adverse drug reaction (from the ADR lexicon) are replaced with \textless MED\textgreater$\ $and \textless ADR\textgreater, respectively (e.g., \textit{\textless MED\textgreater$\ $makes me});
\vspace*{-2mm}
\item Pronoun Lexicon features: the number of tokens from the Pronoun lexicon matched in the tweet;   
\vspace*{-2mm}
\item domain word embeddings: the sum of the domain word embeddings for all tokens in the tweet;
\vspace*{-2mm}
\item domain word clusters: presence of tokens from the domain word clusters. 
\end{itemize}

\textit{Sentiment Lexicon Features} 

We generated features using the sentiment scores provided in the following lexicons: Hu and Liu Lexicon \cite{Hu04}, Norms of Valence, Arousal, and Dominance \cite{warriner2013norms}, labMT \cite{dodds2011temporal}, and NRC Emoticon Lexicon \cite{Kiritchenko2014}. 
The first three lexicons were created through manual annotation while the last one, NRC Emoticon Lexicon, was generated automatically from a large collection of tweets with emoticons. 
The following set of features were calculated separately for each tweet and each lexicon: 
\begin{itemize}
\vspace*{-3mm}
\item  the number of tokens with ${\rm \it score}(w) \neq 0$; 
\vspace*{-1mm}
\item  the total score = $\sum_{w \in {\rm \it tweet}} {\rm \it score}(w)$;
\vspace*{-1mm}
\item  the maximal score = ${\rm \it max}_{w \in {\rm \it tweet}} {\rm \it score}(w)$; 
\vspace*{-1mm}
\item  the score of the last token in the tweet. 
\end{itemize}
\vspace*{-2mm}
We experimented with a number of other existing manually created or automatically generated sentiment and emotion lexicons, such as the NRC Emotion Lexicon \cite{MohammadT13} and the NRC Hashtag Emotion Lexicon \cite{mohammad:2012:STARSEM-SEMEVAL}(http://saifmohammad.com/ WebPages/lexicons.html), but did not observe any improvement in the cross-validation experiments. 
None of the sentiment lexicon features were effective in the cross-validation experiments on Task 1; therefore, we did not include them in the final feature set for this task.

\subsection{Official Submissions}
\label{submissions}

\begin{table*}[t!]
\begin{center}
\begin{tabular}{llccclccc}
\hline {\bf Feature/Parameter}&	& \multicolumn{3}{c}{\bf Task 1 (ADR)} & &\multicolumn{3}{c}{\bf Task 2 (Medication intake)}\\
&	& \multicolumn{3}{c}{submissions} & &\multicolumn{3}{c}{submissions}\\
& & 1 & 2 & 3 & & 1 & 2 & 3\\ 
\hline
\it General textual features & &  & & & \ \ &  & &\\
\ \ \ \ word $n$-grams, $n$ up to & & 3 & 5 & 3 & & 4 & 4 & 4\\
\ \ \ \ non-contiguous $n$-grams, $n$ up to &  & 5 & 3 & 5 & & - & - & -\\
\ \ \ \ character $n$-grams, $n$ up to & & 6 & - & 6 & & 3 & 3 & 3\\
\ \ \ \ unigram stems & & \checkmark & - & \checkmark & & \checkmark & \checkmark & \checkmark\\
\ \ \ \ general-domain word embeddings & & \checkmark & \checkmark & \checkmark & & \checkmark & \checkmark & \checkmark\\
\ \ \ \ general-domain word clusters & & \checkmark & \checkmark & \checkmark & & \checkmark & \checkmark & \checkmark\\
\ \ \ \ negation & & - & - & - & & \checkmark & \checkmark & \checkmark\\
\ \ \ \ Twitter-specific features & & \checkmark & \checkmark & \checkmark & & \checkmark & \checkmark & \checkmark\\
\ \ \ \ punctuation & & \checkmark & \checkmark & \checkmark & &\checkmark & \checkmark & \checkmark\\[3pt]
\it Domain-specific features & & & & & & & &\\
\ \ \ \ domain generalized $n$-grams, $n$ up to & & 4 & 8 & 4 & & 4 & 4 & 4\\
\ \ \ \ domain gen. non-cont. $n$-grams, $n$ up to & & 5 & - & 5 & & 5 & 5 & 5\\
\ \ \ \ ADR lexicon & & \checkmark & \checkmark & \checkmark & & - & \checkmark & - \\
\ \ \ \ Pronoun lexicon & & \checkmark & \checkmark & \checkmark & & - & \checkmark & - \\
\ \ \ \ domain word embeddings & & \checkmark & \checkmark & \checkmark & & \checkmark & \checkmark & \checkmark\\
\ \ \ \ domain word clusters & & \checkmark & \checkmark & \checkmark & & - & - & - \\[3pt]
\it Sentiment lexicon features & & - & - & - & & \checkmark & \checkmark & \checkmark\\[3pt]
\it SVM parameters & & & & & & &  & \\
\ \ \ \ C & & 0.001 & 0.001 & 0.001 & & 0.01 & 0.01 & 0.1\\
\ \ \ \ class weights & & 1, 1 & 1, 1 & 1, 1 & & 4, 2, 1 & 4, 2, 1 & 4, 2, 1\\[3pt]
\it Under-sampling & & & & & & & &\\
\ \ \ \ class proportion & & 1:2 & 1:2 & 1:2, 1:3, 1:4 & & - & - & - \\
\hline
\end{tabular}
\caption{Feature sets and parameters for the three official submissions for Task 1 and Task 2. \checkmark specifies the features included in the classification model; '-' specifies the features not included.}
\label{tab:submission-parameters}
\end{center}
\end{table*}

For each task, our team submitted three sets of predictions. 
The submissions differed in the sets of features and parameters used to train the classification models (Table~\ref{tab:submission-parameters}). 

While developing the system for Task 1 we noticed that the results obtained through cross-validation on the training data were almost 13 percentage points higher than the results obtained by the model trained on the full training set and applied on the development set. 
This drop in performance was mostly due to a drop in precision. 
This suggests that the datasets had substantial differences in the language use, possibly because they were collected and annotated at separate times. 
Therefore, we decided to optimize the parameters and features for submission 1 and submission 2 using two different strategies. 
The models for the three submissions were trained as follows: 
\begin{itemize}
\vspace{-1mm}
\item \textit{Submission 1:} we randomly split the development set into 5 equal folds. 
We trained a classification model on the combination of four folds and the full training set, and tested the model on the remaining fifth fold of the development set. 
The procedure was repeated five times, each time testing on a different fold. 
The feature set and the classification parameters that resulted in the best $F_{class\:1}$ were used to train the final model. 
\vspace{-1mm}
\item \textit{Submission 2:} the features and parameters were selected based on the performance of the model trained on the full training set and tested on the full development set. 
\vspace{-1mm}
\item \textit{Submission 3:} we used the same features and parameters as in submission 1, except we trained an ensemble of three models, varying the class distribution in the sub-sampling procedure (1:2, 1:3, and 1:4). 
\end{itemize}

\vspace{-1mm}
For Task 2, the features and parameters were selected based on the cross-validation results run on the combination of the training and development set. 
We randomly split the development set into 3 equal folds. 
We trained a classification model on the combination of two folds and the full training set, and tested the model on the remaining third fold of the development set. 
The procedure was repeated three times, each time testing on a different fold. 
The models for the three submissions were trained as follows: 
\begin{itemize}
\item \textit{Submission 1:} we used the features and parameters that gave the best results during cross-validation.
\vspace{-1mm}
\item \textit{Submission 2:} we used the same features and parameters as in submission 1, but added features derived from two domain resources: the ADR lexicon and the Pronoun lexicon. 
\vspace{-1mm}
\item \textit{Submission 3:} we used the same features as in submission 1, but changed the SVM C parameter to 0.1. 
\end{itemize}

\vspace{-1mm}
For both tasks and all submissions, the final models were trained on the combination of the full training set and full development set, and applied on the test set. 

\section{Results and Discussion}

{\bf Task 1 (Classification of Tweets for ADR)}

\setlength{\tabcolsep}{12pt}

\begin{table*}[t!]
\begin{center}
\begin{tabular}{lccc}
\hline 	
{\bf Submission} & $P_{class\:1}$ & $R_{class\:1}$ & $F_{class\:1}$ \\ 
\hline
\textit{a. Baselines} &\\
\ \ \ \ \ a.1. Assigning class 1 (ADR) to all instances & 0.077 & 1.000 & 0.143\\
\ \ \ \ \ a.2. SVM-unigrams & 0.391	& 0.298	& 0.339\\[5pt]
\textit{b. Top 3 teams in the shared task} &\\
\ \ \ \ \ b.1. NRC-Canada & 0.392 & 0.488	& 0.435\\
\ \ \ \ \ b.2. AASU & 0.437	& 0.393	& 0.414\\
\ \ \ \ \ b.3. NorthEasternNLP & 0.395	& 0.431	& 0.412\\[5pt]
\textit{c. NRC-Canada official submissions} &\\
\ \ \ \ \ c.1. submission 1 & 0.392 & 0.488	& 0.435 \\
\ \ \ \ \ c.2. submission 2 & 0.386	 & 0.413	& 0.399 \\
\ \ \ \ \ c.3. submission 3 & 0.464	& 0.396	& 0.427 \\[5pt]
\textit{d. Our best result} & 0.398	& 0.508	& 0.446\\
\hline
\end{tabular}
\caption{Task 1: Results for our three official submissions, baselines, and top three teams. Evaluation measures for Task 1 are precision (P), recall (R), and F1-measure (F) for class 1 (ADR). }
\label{tab:task1-submission-results}
\end{center}
\end{table*}

The results for our three official submissions are presented in Table~\ref{tab:task1-submission-results} (rows c.1--c.3). 
The best results in $F_{class\:1}$ were obtained with submission 1 (row c.1). 
The results for submission 2 are the lowest, with F-measure being 3.5 percentage points lower than the result for submission 1 (row c.2). 
The ensemble classifier (submission 3) shows a slightly worse performance than the best result. 
However, in the post-competition experiments, we found that larger ensembles (with 7--11 classifiers, each trained on a random sub-sample of the majority class to reduce class imbalance to 1:2) outperform our best single-classifier model by over one percentage point with $F_{class\:1}$ reaching up to $0.446$ (row d).   
Our best submission is ranked first among the nine teams participated in this task (rows b.1--b.3).

Table~\ref{tab:task1-submission-results} also shows the results for two baseline classifiers. 
The first baseline is a classifier that assigns class 1 (ADR) to all instances (row a.1). 
The performance of this baseline is very low ($F_{class\:1} = 0.143$) due to the small proportion of class 1 instances in the test set. 
The second baseline is an SVM classifier trained only on the unigram features (row a.2). 
Its performance is much higher than the performance of the first baseline, but substantially lower than that of our system. 
By adding a variety of textual and domain-specific features as well as applying under-sampling, we are able to improve the classification performance by almost ten percentage points in F-measure.  

\begin{table*}[t!]
\begin{center}
\begin{tabular}{lccc}
\hline
{\bf Submission} & $P_{class\:1}$ & $R_{class\:1}$ & $F_{class\:1}$\\ 
\hline
a. submission 1 (all features) & 0.392	& 0.488	& 0.435\\[5pt]
b. all $-$ general textual features	& 0.390	& 0.444	& 0.415\\
\ \ \ \ \ b.1. all $-$ general $n$-grams	& 0.397	& 0.484	& 0.436\\
\ \ \ \ \ b.2. all $-$ general embeddings	& 0.365	& 0.480	& 0.414\\
\ \ \ \ \ b.3. all $-$ general clusters	& 0.383	& 0.498	& 0.433\\
\ \ \ \ \ b.4. all $-$ Twitter-specific $-$ punctuation	& 0.382	& 0.494	& 0.431\\[5pt]
c. all $-$ domain-specific features	& 0.341	& 0.523	& 0.413\\
\ \ \ \ \ c.1. all $-$ domain generalized $n$-grams	& 0.366	& 0.514	& 0.427\\
\ \ \ \ \ c.2. all $-$ Pronoun lexicon & 0.385	& 0.496	& 0.433\\
\ \ \ \ \ c.3. all $-$ domain embeddings	& 0.365	& 0.515	& 0.427\\
\ \ \ \ \ c.4. all $-$ domain clusters	& 0.386	& 0.492	& 0.432\\[5pt]
d. all $-$ under-sampling &	0.628	& 0.217	& 0.322\\
\hline
\end{tabular}
\caption{Task 1: Results of our best system (submission 1) on the test set when one of the feature groups is removed.}
\label{tab:task1-ablation-results}
\end{center}
\end{table*}

To investigate the impact of each feature group on the overall performance, we conduct ablation experiments where we repeat the same classification process but remove one feature group at a time. 
Table~\ref{tab:task1-ablation-results} shows the results of these ablation experiments for our best system (submission 1). 
Comparing the two major groups of features, general textual features (row b) and domain-specific features (row c), we observe that they both have a substantial impact on the performance. 
Removing one of these groups leads to a two percentage points drop in $F_{class\:1}$.
The general textual features mostly affect recall of the ADR class (row b) while the domain-specific features impact precision (row c). 
Among the general textual features, the most influential feature is general-domain word embeddings (row b.2). 
Among the domain-specific features, $n$-grams generalized over domain terms (row c.1) and domain word embeddings (row c.3) provide noticeable contribution to the overall performance. 
In the Appendix, we provide a list of top 25 $n$-gram features (including $n$-grams generalized over domain terms) ranked by their importance in separating the two classes.

As mentioned before, the data for Task 1 has high class imbalance, which significantly affects performance.  
Not applying any of the techniques for handling class imbalance, results in a drop of more than ten percentage points in F-measure---the model assigns most of the instances to the majority (non-ADR) class (row d). 
Also, applying under-sampling with the balanced class distribution results in performance significantly worse ($F_{class\:1} = 0.387$) than the performance of the submission 1 where under-sampling with class distribution of 1:2 was applied. 

Error analysis on our best submission showed that there were 395 false negative errors (tweets that report ADRs, but classified as non-ADR) and 582 false positives (non-ADR tweets classified as ADR).
Most of the false negatives were due to the creative ways in which people express themselves (e.g., \textit{i have metformin tummy today :-(} ). 
Large amounts of labeled training data or the use of semi-supervised techniques to take advantage of large unlabeled domain corpora may help improve the detection of ADRs in such tweets.
False positives were caused mostly due to the confusion between ADRs and other relations between a medication and a symptom. 
Tweets may mention both a medication and a symptom, but the symptom may not be an ADR.
The medication may have an unexpected positive effect (e.g., \textit{reversal of hair loss}), or may alleviate an existing health condition. 
Sometimes, the relation between the medication and the symptom is not explicitly mentioned in a tweet, yet an ADR can be inferred by humans. 

\vspace{2mm}
{\bf Task 2 (Classification of Tweets for Medication Intake)}

The results for our three official submissions on Task 2 are presented in Table~\ref{tab:task2-submission-results} (rows c.1--c.3). 
The best results in $F_{class\:1\:+\:class\:2}$ are achieved with submission 1 (row c.1). 
The results for the other two submissions, submission 2 and submission 3, are quite similar to the results of submission 1 in both precision and recall (rows c.2--c.3). 
Adding the features from the ADR lexicon and the Pronoun lexicon did not result in performance improvement on the test set. 
Our best system is ranked third among the nine teams participated in this task (rows b.1--b.3).

\begin{table*}[t!]
\begin{center}
\begin{tabular}{lccc}
\hline
{\bf Submission} & $P_{class\:1\:+\:class\:2}$ & $R_{class\:1\:+\:class\:2}$ & $F_{class\:1\:+\:class\:2}$\\ 
\hline
\textit{a. Baselines} &\\
\ \ \ \ \ a.1. Assigning class 2 to all instances & 0.359 & 0.609 & 0.452\\
\ \ \ \ \ a.2. SVM-unigrams & 0.680	& 0.616	& 0.646\\[5pt]
\textit{b. Top 3 teams in the shared task} &\\
\ \ \ \ \ b.1. InfyNLP & 0.725	& 0.664	& 0.693\\
\ \ \ \ \ b.2. UKNLP & 0.701 &	0.677	& 0.689\\
\ \ \ \ \ b.3. NRC-Canada & 0.708	& 0.642	& 0.673\\[5pt]
\textit{c. NRC-Canada official submissions} &\\
\ \ \ \ \ c.1. submission 1 & 0.708	& 0.642	& 0.673\\
\ \ \ \ \ c.2. submission 2 & 0.705	& 0.639	& 0.671\\
\ \ \ \ \ c.3. submission 3 & 0.704	& 0.635	& 0.668\\
\hline
\end{tabular}
\caption{Task 2: Results for our three official submissions, baselines, and top three teams. Evaluation measures for Task 2 are micro-averaged P, R, and F1-score for class 1 (intake) and class 2 (possible intake).}
\label{tab:task2-submission-results}
\end{center}
\end{table*}

\setlength{\tabcolsep}{6pt}

\begin{table*}[t!]
\begin{center}
\begin{tabular}{lccc}
\hline
{\bf Submission} & $P_{class\:1\:+\:class\:2}$ & $R_{class\:1\:+\:class\:2}$ & $F_{class\:1\:+\:class\:2}$\\ 
\hline
a. submission 1 (all features) & 0.708	& 0.642	& 0.673\\[5pt]
b. all $-$ general textual features	& 0.697	& 0.603	& 0.647\\
\ \ \ \ \ b.1. all $-$ general $n$-grams	& 0.676	& 0.673	& 0.674\\
\ \ \ \ \ b.2. all $-$ general embeddings	& 0.709	& 0.638	& 0.671\\
\ \ \ \ \ b.3. all $-$ general clusters	& 0.685	& 0.671	& 0.678\\
\ \ \ \ \ b.4. all $-$ negation $-$ Twitter-specific $-$ punctuation	& 0.683	& 0.670	& 0.676\\[5pt]
c. all $-$ domain-specific features	& 0.679	& 0.653	& 0.666\\
\ \ \ \ \ c.1. all $-$ domain generalized $n$-grams	& 0.680	& 0.652	& 0.665\\
\ \ \ \ \ c.2. all $-$ domain embeddings	& 0.682	& 0.671	& 0.676\\[5pt]
d. all $-$ sentiment lexicon features	& 0.685	& 0.673	& 0.679\\[5pt]
e. all $-$ class weights	& 0.718	& 0.645	& 0.680\\
\hline
\end{tabular}
\caption{Task 2: Results of our best system (submission 1) on the test set when one of the feature groups is removed.}
\label{tab:task2-ablation-results}
\end{center}
\end{table*}

Table~\ref{tab:task2-submission-results} also shows the results for two baseline classifiers. 
The first baseline is a classifier that assigns class 2 (possible medication intake) to all instances (row a.1). 
Class 2 is the majority class among the two positive classes, class 1 and class 2, in the training set. 
The performance of this baseline is quite low ($F_{class\:1\:+\:class\:2} = 0.452$) since class 2 covers only 36\% of the instances in the test set. 
The second baseline is an SVM classifier trained only on the unigram features (row a.2). 
The performance of such a simple model is surprisingly high ($F_{class\:1\:+\:class\:2} = 0.646$), only 4.7 percentage points below the top result in the competition. 

Table~\ref{tab:task2-ablation-results} shows the performance of our best system (submission 1) when one of the feature groups is removed. 
In this task, the general textual features (row b) played a bigger role in the overall performance than the domain-specific (row c) or sentiment lexicon (row d) features. 
Removing this group of features results in more than 2.5 percentage points drop in the F-measure affecting both precision and recall (row b). 
However, removing any one feature subgroup in this group (e.g., general $n$-grams, general clusters, general embeddings, etc.) results only in slight drop or even increase in the performance (rows b.1--b.4). 
This indicates that the features in this group capture similar information. 
Among the domain-specific features, the $n$-grams generalized over domain terms are the most useful. 
The model trained without these $n$-grams features performs almost one percentage point worse than the model that uses all the features (row c.1). 
The sentiment lexicon features were not helpful (row d). 

Our strategy of handling class imbalance through class weights did not prove successful on the test set (even though it resulted in increase of one point in F-measure in the cross-validation experiments). 
The model trained with the default class weights of 1 for all classes performs 0.7 percentage points better than the model trained with the class weights selected in cross-validation (row e). 

The difference in how people can express medication intake vs.\@ how they express that they have not taken a medication can be rather subtle.
For example, the expression \textit{I need Tylenol} indicates that the person has not taken the medication yet (class 3), whereas the expression \textit{I need more Tylenol} indicates that the person has taken the medication (class 1). 
In still other instances, the word \textit{more} might not be the deciding factor in whether a medication was taken or not (e.g., \textit{more Tylenol didn't help}). A useful avenue of future work is to explore the role function words play in determining the semantics of a sentence, specifically, when they imply medication intake, when they imply the lack of medication intake, and when they are not relevant to determining medication intake.

\section{Conclusion}

Our submissions to the 2017 SMM4H Shared Tasks Workshop obtained the first and third ranks in Task1 and Task 2, respectively. 
In Task 1, the systems had to determine whether a given tweet mentions an adverse drug reaction. 
In Task 2, the goal was to label a given tweet with one of the three classes: personal medication intake, possible medication intake, or non-intake. 
For both tasks, we trained an SVM classifier leveraging a number of textual, sentiment, and domain-specific features. 
Our post-competition experiments demonstrate that the most influential features in our system for Task 1 were general-domain word embeddings, domain-specific word embeddings, and $n$-grams generalized over domain terms. 
Moreover, under-sampling the majority class (non-ADR) to reduce class imbalance to 1:2 proved crucial to the success of our submission. 
Similarly, $n$-grams generalized over domain terms improved results significantly in Task 2. 
On the other hand, sentiment lexicon features were not helpful in both tasks.
\makeatletter
\renewcommand{\@biblabel}[1]{\hfill #1.}
\makeatother

\setlength{\bibitemsep}{.2\baselineskip}

\bibliography{adr}
\bibliographystyle{unsrt}

\section*{Appendix}

We list the top 25 $n$-gram features (word $n$-grams and $n$-grams generalized over domain terms) ranked by mutual information of the presence/absence of $n$-gram features ($f$) and class labels ($C$):
\[
I(f, C) = \sum \nolimits_{c \in C} \sum\nolimits_{f \in \{present,absent\}} p(f,c)\;log \left(\frac{p(f,c)}{p(f)\;p(c)}\right),
\]
where $C = \{0,1\}$ for Task 1 and $C = \{1, 2, 3\}$ for Task 2. 

Here, \textless ADR\textgreater$\ $represents a word or a phrase from the ADR lexicon; \textless MED\textgreater$\ $ represents a medication name from our one-word medication list.

\begin{multicols}{4}
\textbf{Task 1}

1. me\\
2. withdraw\\
3. i\\
4. makes\\
5. \textless ADR\textgreater$\ $.\\
6. makes me\\
7. feel\\
8. me \textless ADR\textgreater\\
9. \textless MED\textgreater$\ $\textless ADR\textgreater\\
10. made me\\
11. withdrawal\\
12. \textless MED\textgreater$\ $makes\\
13. my\\
\columnbreak

$\ $

14. \textless MED\textgreater$\ $makes me\\
15. gain\\
16. weight\\
17. \textless ADR\textgreater$\ $and\\
18. headache\\
19. made\\
20. tired\\
21. rivaroxaban diary\\
22. withdrawals\\
23. zomby\\
24. day\\
25. \textless MED\textgreater$\ $diary\\
\columnbreak

\textbf{Task 2}

1. steroids\\
2. need\\
3. i need\\
4. took\\
5. on steroids\\
6. on \textless MED\textgreater\\
7. i\\
8. i took\\
9. http://someurl\\
10. @username\\
11. her\\
12. on\\
13. him\\
\columnbreak

$\ $

14. you\\
15. he\\
16. me\\
17. need a\\
18. kick\\
19. i need a\\
20. she\\
21. headache\\
22. kick in\\
23. this \textless MED\textgreater\\
24. need a \textless MED\textgreater\\
25. need \textless MED\textgreater\\
\end{multicols}

\end{document}